\documentclass[journal,twoside,web]{ieeecolor}
\usepackage[T1]{fontenc}
\usepackage{generic}
\usepackage{amsmath,amssymb,amsfonts}
\usepackage{algorithm}
\usepackage{algpseudocode}
\usepackage{graphicx}
\usepackage{hyperref}
\hypersetup{hidelinks=true}
\usepackage{textcomp}
\usepackage{booktabs}
\usepackage{caption}
\usepackage{float}

\usepackage[backend=biber,natbib=true,maxcitenames=2,mincitenames=1]{biblatex}

\begin{document}

\title{FoundationalECGNet: A Lightweight Foundational Model for ECG-based Multitask Cardiac Analysis}
\author{Md. Sajeebul Islam Sk.\textsuperscript{1}, Md Jobayer\textsuperscript{1}, Md Mehedi Hasan Shawon\textsuperscript{1} and Md. Golam Raibul Alam
\thanks{This work is supported by the Research Seed Grant Initiative-2024, BRAC University, Dhaka-1212, Bangladesh.}
\thanks{\textsuperscript{1}These authors contributed equally to this work.}
\thanks{Md. Sajeebul Islam Sk. and Md. Golam Raibul Alam are with the Department of Computer Science and Engineering, BRAC University, Dhaka 1212, Bangladesh (e-mail: mdsajeebulislamsk@gmail.com, rabiul.alam@bracu.ac.bd).}
\thanks{Md Jobayer and Md Mehedi Hasan Shawon are with the Department of Electrical and Electronic Engineering, BRAC University, Dhaka 1212, Bangladesh (e-mail: \{jobayer, mehedi.shawon\}@ieee.org ).}
\thanks{Md Jobayer is also with the Department of Biomedical Engineering, Linköping University, Linköping 58183, Sweden.}
}

\maketitle
\begin{abstract}
Cardiovascular diseases (CVDs) remain a leading cause of mortality worldwide, underscoring the importance of accurate and scalable diagnostic systems. Electrocardiogram (ECG) analysis is central to detecting cardiac abnormalities, yet challenges such as noise, class imbalance, and dataset heterogeneity limit current methods. To address these issues, we propose FoundationalECGNet, a foundational framework for automated ECG classification. The model integrates a dual-stage denoising by Morlet and Daubechies wavelets transformation, Convolutional Block Attention Module (CBAM), Graph Attention Networks (GAT), and Time Series Transformers (TST) to jointly capture spatial and temporal dependencies in multi-channel ECG signals. FoundationalECGNet first distinguishes between Normal and Abnormal ECG signals, and then classifies the Abnormal signals into one of five cardiac conditions: Arrhythmias, Conduction Disorders, Myocardial Infarction, QT Abnormalities, or Hypertrophy. Across multiple datasets, the model achieves a 99\% F1-score for Normal vs. Abnormal classification and shows state-of-the-art performance in multi-class disease detection, including a 99\% F1-score for Conduction Disorders and Hypertrophy, as well as a 98.9\% F1-score for Arrhythmias. Additionally, the model provides risk level estimations to facilitate clinical decision-making. In conclusion, FoundationalECGNet represents a scalable, interpretable, and generalizable solution for automated ECG analysis, with the potential to improve diagnostic precision and patient outcomes in healthcare settings. We'll share the code after acceptance.
\end{abstract}

\section{Introduction}
\label{sec:introduction}
\IEEEPARstart{C}{ardiovascular} diseases (CVDs) are the leading cause of death worldwide, leading to approximately 17.9 million deaths each year \citep{csahin2022risk}. Early and accurate diagnosis of these diseases is crucial to reduce the number of deaths, as timely intervention can significantly improve patient outcomes \citep{almansouri2024early}. Electrocardiography (ECG) is a common, non-invasive, and low-cost test used to find problems such as arrhythmias and heart attacks \citep{gunasekaran2024artifact}. 

The ECG signal is inherently non-stationary, with its frequency content varying over time, and it is frequently affected by artifacts such as baseline wander from patient movement, power-line interference from electrical sources, and electromyographic noise from muscle contractions or electrode motion \citep{A_NG}.
Signals frequently contain interference from muscle activity, fluctuations, electrode displacement, and power line disturbances, which obscure crucial patterns, particularly in portable and wearable devices \citep{zhang2025opportunities}. Additionally, the complexity of ECG data arises from its multiple channels, and the difficulty to detect minor changes in the ECG signals \citep{AYYUB2025110594}.
Moreover, manually annotating ECG signals is labor-intensive that also varies from one annotator to another, resulting in errors and making it clinically unreliable \citep{wang2021interactive}.

Deep learning methods have emerged as viable alternatives that perform automatic feature extraction and leveraging clinical datasets to enhance performance in ECG classification tasks \citep{butt2022toward}. Among others, Convolutional Neural Networks (CNNs) excels at identifying spatial patterns in ECG signals, enabling accurate detection of arrhythmias without manual intervention \citep{kim2025novel}. Transformer-based approaches advance this by capturing long-range dependencies, which aids in identifying subtle morphological changes associated with conditions such as myocardial infarctions. However, among the presence of other limitations such as interpretability and computational efficiency, generalization remains a difficult tasks for these models \citep{atwa2025interpretable}. 

Noise creates additional difficulties, diminishing the signal quality and their utility in clinical environments \citep{rahman2024robustness}. Additionally, class imbalance issue leads to biased predictions favoring prevalent classes \citep{monachino2023deep}. The scarcity of high-quality, annotated ECG datasets further presents a major obstacle, as it limits the diversity and amount of training data available for model development \citep{lai2025diffusets}.

To address these research gaps, it is required to develop a versatile ECG model that integrates multiple capabilities. This includes learning from diverse data sources including multi-channel data and devices, employing effective denoising techniques that preserve clinically relevant features, modeling relationship between the ECG leads and temporal dependencies, and ensuring robustness in the presence of class imbalance. The model minimizes computational requirements through efficient architectures, delivers more reliable predictions, reduces overfitting, enhances generalization across diverse clinical contexts for reliable clinical decision-making. 

In this way, we propose a foundational model called FoundationalECGNet that incorporates wavelet-based denoising, graph attention networks, and transformer networks that mitigates the above-mentioned limitations. Our main objective is to improve the reliability and accessibility of automated ECG interpretation, enabling earlier detection of cardiovascular disorders and enhancing decision support for healthcare professionals, with the following key contributions:
\begin{itemize}
\item We propose FoundationalECGNet, a foundational model framework for classifying ECG signals which is pre-trained on diverse datasets. It captures both local and global features of ECG signals and learns intrinsic patterns from different sources, enabling it to adapt well to various multi-channel ECG datasets.
\item We employ a dual-stage noise removal technique utilizing Morlet and Daubechies (DB4) at order 4 wavelets to denoise ECG signals effectively. This process eliminates interference while retaining vital signal characteristics by breaking down the data into time-frequency elements, enabling precise noise suppression.
\item Our proposed method integrates Graph Attention Networks (GAT), Time Series Transformers (TST), and the Convolutional Block Attention Module (CBAM) to learn relevant features from ECG signals simultaneously. GAT focuses on relationships across channels whereas TST captures extended temporal patterns for subtle anomaly detection, and CBAM detects critical spatial and channel-specific elements.
\item We utilize multi-level ECG datasets to initially classify signals as Normal or Abnormal. The model performs an extensive analysis of abnormal signals to determine the most likely specific disease. In order to assist with decision-making, the predictions are supplemented by a disease probability score and a corresponding clinical report.
\item Our proposed model is computationally efficient compared to existing foundational models and other ECG-based state-of-the-art models. The proposed model contains approximately 7.5 million parameters, making it two times smaller than one of the existing models.

\end{itemize}
The paper is organized as follows: in Section II, we discuss previous work. Section III goes into detail about the proposed FoundationalECGNet model. Section IV describes the experimental setup and the datasets. Results are provided in Section V. Section VI concludes the paper by summarizing the main findings of FoundationalECGNet and suggesting future directions for research.

\section{RELATED WORK}
ECG signals from wearable devices often contain noise so effective denoising is essential to preserve key features for diagnosis \citep{LOMOIO2025103058}. Traditional methods like bandpass filtering are common but often fail to retain subtle signal details. Wavelet based denoising using Symlet and Daubechies (DB2) wavelets effectively removes noise while preserving critical ECG features in low signal-to-noise ratio \citep{yadav2016denoising}. Deep learning-based denoising has also shown promise. Autoencoder models trained on synthetic noise-augmented ECGs demonstrate robustness to real-world noise, with average improved SNR of 27.45 dB for baseline wander, 25.72 dB for muscle artifacts, and 29.91 dB for electrode motion artifacts \citep{electronics12071606}. However, such methods often require paired noisy and clean data, which may not be readily available in clinical practice. Hybrid approaches combining wavelet based and deep learning techniques are developed for balancing computational efficiency and denoising performance \citep{balasubramaniam2009implementation}.

Class imbalance is a major challenge in ECG datasets where normal recordings outnumber pathological ones that leading to biased predictions \citep{S2021102779}. The ADASYN method generates synthetic samples for minority classes, improving performance on rare conditions like ventricular tachycardia with F-measures such as 0.8505 on the Vehicle dataset and 0.5726 on the Pima Indian Diabetes dataset \citep{alhudhaif2021novel}. Recent studies have integrated ADASYN with deep learning frameworks to enhance classification accuracy in imbalanced datasets, achieving an average AUC of 0.902 and Fmax of 0.749 with bidirectional LSTM models \citep{9190034}. Dataset heterogeneity, arising from variations in devices, sampling rates, and patient demographics, further complicates model generalization \citep{gliner2023using}. Public datasets like PTB-XL \citep{PTB-XL} and PhysioNet/CinC \citep{AF} Challenge 2017 provide standardized benchmarks for ECG research. However, most models are trained on single datasets, limiting their applicability in diverse clinical scenarios.

The automatic classification of electrocardiogram (ECG) signals has improved significantly with the use of deep learning models, including convolutional neural networks (CNNs) and transformer-based systems \citep{Wu}. These approaches help manage the complex nature of ECG signals, leading to better accuracy in identifying heart diseases and classifying arrhythmias \citep{GUPTA2024e26787}. However, challenges remain in making these systems more robust and able to work well in real clinical settings \citep{silva2025systematicreviewecgarrhythmia}.
Early methods for classifying ECG signals relied on handcrafted features, which required extensive expert knowledge \citep{PTB-XL}. Features like heart rate, QRS duration, and morphological aspects such as R-wave amplitude were extracted manually to detect conditions like arrhythmias or myocardial infarction \citep{PTB-XL}. These techniques performed adequately in controlled environments but struggled with the variability of ECG signals across different patients and devices \citep{lahmiri2014comparative}. Manual feature extraction \citep{AZZOUZ2024e26171} was time-consuming and prone to errors due to subjective interpretation particularly on a large scale. These limitations highlighted the need for automated methods that can scale and handle natural differences in ECG data.
Deep learning has transformed ECG analysis by automating feature extraction and increasing classification accuracy for various heart conditions \citep{khalid2024applications}. CNNs are widely used because they effectively capture spatial and temporal patterns in ECG signals \citep{hannun2019cardiologist}. CNN model trained on a proprietary dataset of over 50,000 patients achieved an average F1 score of 0.837 and AUC of 0.97 in detecting arrhythmias from single-lead ECGs \citep{hannun2019cardiologist}. Similarly, CNN architectures with residual connections have shown strong performance on the PhysioNet/CinC Challenge 2017 dataset \citep{AF} for multi-lead ECG classification \citep{ansari2023deep}. However, these models are often optimized for specific datasets, limiting their use in other contexts \citep{weimann2021transfer}. Transformer based architectures have gained attention for modeling long-range temporal dependencies in ECG signals, outperforming CNNs in detecting subtle changes associated with myocardial infarction \citep{HU2022105325}. Still, transformers require substantial computational resources and large training datasets, which can be challenging in resource limited clinical settings\citep{zhao2023transformingecgdiagnosisanindepth}. Graph Attention Networks (GATs) \citep{GAT} have been explored to capture inter lead relationships in multi-lead ECGs. GAT based frameworks have improved detection of spatially distributed abnormalities such as those in ischemic heart disease with accuracy reaching 98.50\% and F1 scores around 98.71\% on related tests though evaluations were limited to single datasets \citep{DUONG2023120107}.

Transfer learning and domain adaptation have been explored to address this issue \citep{weimann2021transfer}. For instance, pre-training on diverse datasets followed by fine-tuning on target datasets can mitigate domain shift, enhancing model robustness across different acquisition conditions, with AUC of 0.90 for heart failure prediction \citep{10.1371/journal.pone.0316043}.
Foundation models, pre-trained on mixed datasets and adaptable to multiple tasks, have become prominent in fields like medical imaging and natural language processing, but they remain underexplored in ECG analysis \citep{han2024foundationmodelselectrocardiogramreview}. A foundation ECG model recently proposed was pre-trained on public and private datasets, achieving strong performance in arrhythmia detection tasks and heart failure prediction, with an average improvement of 6\% across four key tasks \citep{song2025cremacontrastiveregularizedmasked}. However, reliance on private data limits reproducibility \citep{large-scale-training}. 

Conversely, our proposed FoundationalECGNet utilizes open-access datasets, integrates advanced multiple wavelet-based denoising, and utilizes GAT \citep{GAT} and transformer-based architectures to extract and learn multidimensional ECG features. We intend to bridge task-specific models and clinical deployment by correcting class imbalance with ADASYN and pre-training on diverse datasets.

\section{METHODOLOGY}
\begin{figure*}[htbp]
\centering
    \includegraphics[width=0.70\linewidth]{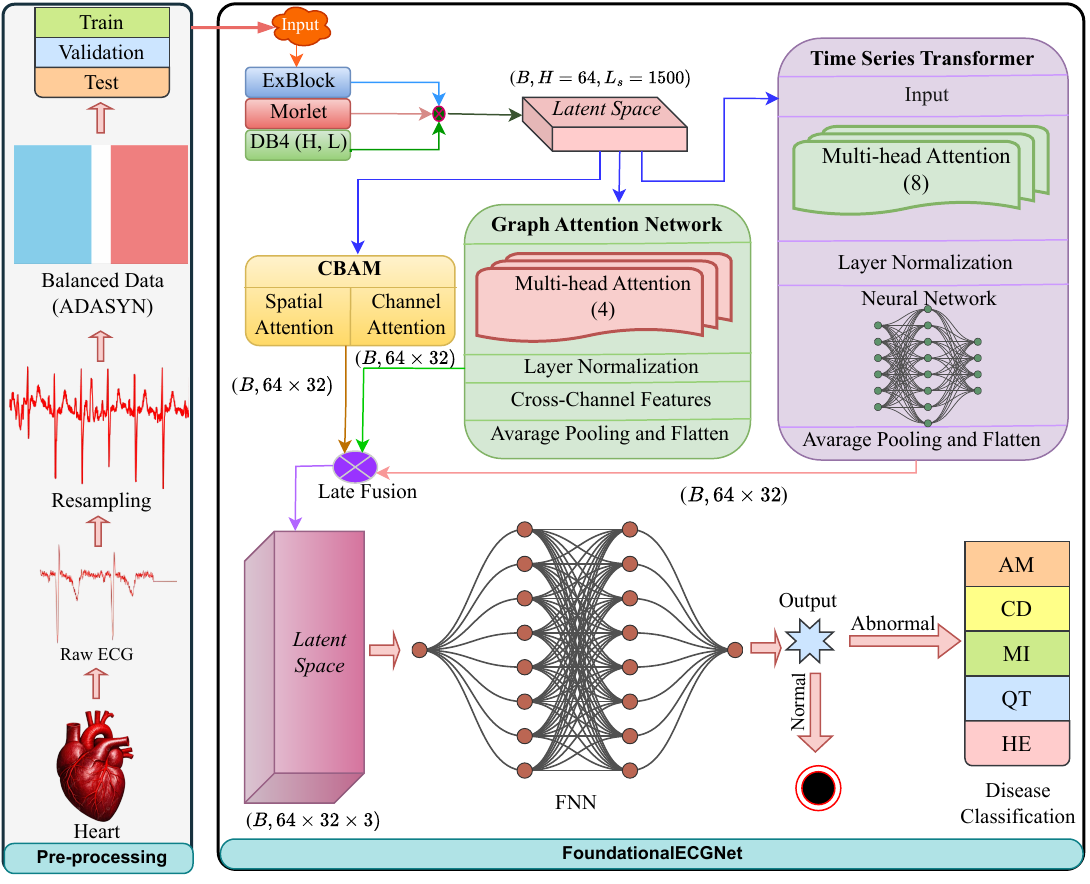}
\caption{Overview of the FoundationalECGNet architecture, consisting of Disease Classification. The model uses Graph Attention Networks and Multi-head Attention for effective feature extraction, with Late Fusion applied for final decision-making.}
\label{fig:0}
\end{figure*}

We show the overall framework of our proposed method in Figure \ref{fig:0}, which comprises two main parts: (a) data preprocessing, (b) model architecture. 
The model initially categorizes ECG signals as Normal or Abnormal, and if identified as Abnormal, it further specifies conditions: Arrhythmia (AM), Conduction Disorders (CD), Myocardial Infarction (MI), QT Prolongation (QT), or Hypertrophy and Enlargement (HE). The details are explained below.

\subsection{Pre-Processing}
We resample each signal to 1500 samples \citep{zhang2023brant} to ensure consistency throughout the dataset. Additionally, the signals are processed to a 250Hz sample rate, standardizing the temporal resolution for consistent feature extraction \citep{zhang2023brant}. As the ECG data comes from multiple channels, we process each channel independently during pre-processing. We use Min-Max Normalization to each signal on an individual channel basis, fixing each to a range of 0 to 1. This ensures the elimination of bias resulting from different magnitudes across channels in the model's learning process. Subsequent to normalization, the data is transformed into a 2D array, making it appropriate for model input. In ECG data, the majority is normal, while abnormal instances are sparse \citep{wang2024anyecg}. Basic techniques such as random oversampling or SMOTE can balance the dataset but frequently introduce several overlapping or less useful samples \citep{mujahid2024data}. We select ADASYN because of its ability to generate new samples particularly for challenging and unclear irregular data, also improving the model's focus on rare but important instances, resulting in improved cardiac identification compared to different approaches \citep{mavaddati2025ecg}.
\subsection{Model Architecture}
Our proposed FoundationalECGNet model shown in Figure \ref{fig:0}, processes the raw ECG data to extract features in eq. \ref{eq:morlet}- \ref{eq:latent}, apply attention mechanisms and classify. In this process, ECG data is passed through the feature extraction block where a number of transformations are applied to improve the signal quality and extract important characteristics to classify. Morlet wavelet reduces initial noise and extracts QRS complex features using the Gaussian function \citep{lin2000feature}, whereas Daubechies wavelets of order 4 (DB4) \citep{db4} mitigate additional noise and identify R peaks from the ECG. Both methods are applied to input ECG signals to configure the model for time-frequency and temporal representation. The Morlet is expressed as:
\begin{equation}
\mathbf{F_{\text{Morlet}}}(t) = \exp\left(-\frac{t^2}{2\sigma^2}\right) \cos(5t)
\label{eq:morlet}
\end{equation}
In equation \ref{eq:morlet}, where
    \( t \) represent time,
    \( \sigma \) controls the width of ECG, and the factor 5 is chosen to capture the frequency range corresponding to the QRS complex features.
The DB4 wavelet is often used for complex transformation of ECG signals through analysis of low-frequency baseline changes and high-frequency structural details \citep{db4}. The low-frequency component is extracted using the following scaling function:
\begin{equation}
\mathbf{f_{\text{low}}}(t) = h_0 \cdot \phi(t) + h_1 \cdot \phi(t-1) + h_2 \cdot \phi(t-2) + h_3 \cdot \phi(t-3)
\label{eq:low}
\end{equation}
\begin{equation}
\mathbf{f_{\text{high}}}(t) = g_0 \cdot \psi(t) + g_1 \cdot \psi(t-1) + g_2 \cdot \psi(t-2) + g_3 \cdot \psi(t-3)
\label{eq:high}
\end{equation}

In equation \ref{eq:low}, and \ref{eq:high}, the low-pass coefficients 
\( h_0, h_1, h_2, h_3 \) extract the baseline and P–T wave components, 
while the high-pass coefficients \( g_0, g_1, g_2, g_3 \) capture sharp QRS complex features respectively \citep{db4} \citep{liu1991multi}. Finally, we combine the high and low-frequency components for future propagation, which is defined as:
\begin{equation}
    \mathbf{F_{DB4}} = \mathbf{f_{\text{high}}}(t) \cdot \mathbf{f_{\text{low}}}(t)
\label{eq:db4}
\end{equation}
Thus, Wavelets show limited adaptability to various ECG patterns \citep{stkepien2010investigating}, linear decomposition poorly illustrates non-linear relationships in complex signals \citep{liu1991multi}.

The Extended Block (ExBlock) \citep{exblock} is formed with two CBR (Convolution–BatchNorm–ReLU) \citep{cbr} modules that mitigate wavelet limitations by extracting non-linear complex features. Each CBR module employs 1D convolution for feature extraction, batch normalization for stability, and ReLU for non-linearity \citep{cbr}.From \citep{cbr}, the CBR operation can be defined as:
\begin{equation}
\mathbf{F}_{\text{ExBlock}} = \text{ReLU} \left( \text{BN} \left( \mathbf{W} * \mathbf{X}_{\text{in}} + \mathbf{b} \right) \right)
\label{eq:cbr}
\end{equation}
In equation \ref{eq:cbr}, \(\mathbf{X}_{\text{in}}\) is the input tensor with shape (N, H, L), where N is the batch size, H is hidden channels (64), and L is the signal length (1500), \(\mathbf{W}\) and \(\mathbf{b}\) are the
convolutional weights and bias respectively, \(\mathbf{BN}\) denotes batch normalization,
and ReLU is the activation function. The features extracted from ExBlock,
Morlet, and DB4 are concatenated into a latent representation:
\begin{equation}
\mathbf{F{\text{latent}}} = \text{Concat}(\mathbf{F_{\text{ExBlock}}},
\mathbf{F_{\text{Morlet}}}, \mathbf{F_{\text{DB4}}})
\label{eq:latent}
\end{equation}
where \(\mathbf{F_{\text{latent}}}\) provides a comprehensive feature set
of the ECG signal for downstream analysis.
Then, we use Graph Attention Networks (GAT) \citep{GAT} and Convolutional Block Attention Module (CBAM) \cite{exblock} to learn the most important ECG features by prioritizing relevant channels and spatial regions.
The Graph Attention Network (GAT) is used to model the relationships between the different channels in the ECG signal \citep{GAT}. It computes attention scores for each channel pair \( i \) and \( j \) using self-attention \citep{GAT}:
\begin{equation}
a_{ij} = \frac{\exp\left(\text{LeakyReLU}\left(\mathbf{a}^T [W\mathbf{h}_i || W\mathbf{h}_j]\right)\right)}{\sum_{k \in \mathcal{N}_i} \exp\left(\text{LeakyReLU}\left(\mathbf{a}^T [W\mathbf{h}_i || W\mathbf{h}_k]\right)\right)}
\label{eq:GAT}
\end{equation}
In equation \ref{eq:GAT}, \( \mathbf{h}_i \) and \( \mathbf{h}_j \) are the feature vectors for channels \( i \) and \( j \), \( W \) is the weight matrix for channel features, \( \mathcal{N}_i \) is the set of neighboring channels for channel \( i \), \( \mathbf{a} \) is a learnable attention vector. This vector enhances abnormality detection by focusing on critical channel dependencies.

The Convolutional Block Attention Module leverages two attention mechanisms to refine feature learning for applications analyzing complex signals like electrocardiograms. The channel Attention is computed by first applying average and max pooling spatially, passing the results through fully connected layers, and applying a sigmoid activation. Spatial Attention determines crucial regions in the feature map by pooling across channels to obtain 1D feature descriptors, concatenating the pooled representations, and applying a 1D convolution to extract spatial dependencies. The spatial attention map is calculated by averaging and maximizing across channels, concatenating the results, and applying a convolutional layer.

The Convolutional Block Attention Module (CBAM) enhances feature learning by emphasizing significant channels and spatial areas in ECG signal analysis. It utilizes two attention mechanisms: Channel Attention and Spatial Attention \citep{exblock}. The channel Attention is computed by first applying average and max pooling spatially, passing the results through fully connected layers, and applying a sigmoid activation, whereas spatial attention mechanism identifies important regions in the feature map by pooling along the channel dimension \citep{exblock}. The channel and spatial attention define as : 
\begin{equation}
\mathbf{a}_c = \sigma( \text{FC} \left( \text{AP}(x) + \text{MP}(x) \right) )
\label{eq:cbam1}
\end{equation}
\begin{equation}
\mathbf{a}_s = \sigma ( \text{Conv1d}(Concat(AP(x), MP(x)))
\label{eq:cbam}  
\end{equation}

In \ref{eq:cbam1} and \ref{eq:cbam}, \(\mathbf{a}_c\) and \(\mathbf{a}_s\) represent as Channel attention and spatial attention respectively, \( \text{AP}(x) \) and \( \text{MP}(x) \) refer to the AvgPool and MaxPool operations in spatial dimensions, \( \text{FC}_1 \) and \( \text{FC}_2 \) signify fully connected layers, and \( \sigma \) denotes the sigmoid activation function. Then, the final output feature map is generated by concatenating both channel and spatial attention output to the input feature map (x):
\begin{equation}
x_{\text{CBAM}} = x \cdot \mathbf{a}_c \cdot \mathbf{a}_s  
\end{equation}
One potential shortcoming of CBAM module is it can't capture long term dependencies \citep{exblock}.
The Time Series Transformer (TST) is designed to capture long range dependencies within the ECG signal \citep{HU2022105325}. By using multi-head attention the transformer learns global temporal patterns which are crucial for detecting abnormalities. From \citep{sk2024unveiling}, we can be defined as:
\begin{equation}
    \text{Attention}(Q, K, V) = \text{softmax}\left(\frac{QK^T}{\sqrt{d_k}}\right) V
\label{tm}
\end{equation}
In equation \ref{tm} parameters are focusing on long term dependent features in the signal. 

After applying the attention mechanisms and transformer, the features from the GAT, CBAM, and TST are fused into a single latent space representation. 

\begin{equation}
\text{Latent Features} = \text{Concat}(x_{\text{GAT}}, x_{\text{CBAM}}, x_{\text{Transformer}})
\label{ls}
\end{equation}
In equation \ref{ls}, these latent features are passed through an adaptive pooling layer to compress the temporal dimension, followed by fully connected layers to produce the final classification output. The final classification is computed as:
\begin{equation}
  y = \text{ReLU}(W_1 x + b_1) W_2 + b_2
\label{ls2}
\end{equation}
In equation \ref{ls2}, \( W_1 \) and \( W_2 \) are the weight matrices, \( b_1 \) and \( b_2 \) are the bias terms, \( x \) is the input feature vector from the latent space. After obtaining the latent features, these are passed through the classification head, where they are processed by fully connected layers to predict whether the ECG is Normal or Abnormal. 

The detailed workflow of the model can be summarized in algorithm \ref{algo}.
\begin{algorithm}[htbp]
\caption{Fixed FoundationalECGNet Architecture}
\begin{algorithmic}[1]
    \State \textbf{Input:} Raw ECG data in the form of multi-channel time-series signals
    \State \textbf{Output:} Binary or multi-class prediction (Normal or Abnormal, and if Abnormal: Heart Condition)
    \State $A \gets \emptyset$
    \For{\textbf{each} \textit{path} in \textit{ECG signal paths}}
        \State \textit{signal} $\gets$ \texttt{readECGData(path)}
        \State \textit{signal} $\gets$ \texttt{resample(signal, 250Hz)}
        \State $A \gets A \cup \{\textit{signal}\}$
    \EndFor
    \State $A \gets$ \texttt{normalize(A)}
    \State $A \gets$ \texttt{reshape(A, (1500 * n\_channels))}
    \State $A \gets$ \texttt{applyWaveletTransforms(A)}
    \State $A \gets$ \texttt{denoise(A)}
    \State $C \gets \emptyset$
    \For{\textbf{each} \textit{signal} in $A$}
        \State \textit{features} $\gets$ \texttt{extractFeatures(signal)}
        \State $C \gets C \cup \{\textit{features}\}$
    \EndFor
    \State $C \gets$ \texttt{adaptivePooling(C)}
    \State $C \gets$ \texttt{flatten(C)}
    \State $Y \gets$ \texttt{ADASYN(C, Y)}
    \State $X_{train}, X_{val}, Y_{train}, Y_{val} \gets$ \texttt{trainValSplit(C, Y)}
    \State Train model $M \gets$ \texttt{FoundationalECGNet()}
    \State $X_{train}, X_{test}, Y_{train}, Y_{test} \gets$ \texttt{trainTestSplit(C, Y, test\_size=0.2)}
    \State \texttt{output} $\gets M.\texttt{predict}(X_{test})$
\end{algorithmic}
\label{algo}
\end{algorithm}

We develop a neuro-symbolic framework in our proposed FoundationalECGNet for precise ECG disease classification. It combines neural networks for pattern recognition with symbolic reasoning for interpretable, rule-based disease labelling \cite{martinez2023current}. 

After detecting abnormalities, the symbolic module extracts clinical information from the ECG data and calculates disease probability using predefined thresholds. 

From \cite{boulif2023literature}, ECG signal $s(t)$ at $f_s = 250$ Hz, R-peaks are detected as:
\begin{equation}
R = \{ t \mid s(t) > 0.3 \cdot \max(s), \, \text{dist} \geq 0.4 f_s \}
\label{eq:rpeaks}
\end{equation}

This identifies the QRS complex, RR intervals, ST elevation for heart rate and interval calculations \cite{boulif2023literature}. Additional peaks and points are situated inside time periods dependent on $f_s$.
For arrhythmias \cite{boulif2023literature}:
\begin{equation}
P_{\text{AM}} = \max(0.1, \min(0.9, |\text{HR} - 75|/50)),
\label{eq:parr1}
\end{equation}

For conduction disorders (PR $\textgreater$ 200 ms, QRS $\textgreater$ 120 ms \cite{de2008basic}):
\begin{equation}
P_{\text{CD}} = \min\left(0.9, 0.1 + \sum_{i=\text{P}, \text{Q}} \min\left(0.4, \frac{\Delta_i - \text{O}_i}{\text{S}_i}\right) \right)
\label{eq:pcond}
\end{equation}
For myocardial infarction (ST $\textgreater$ 0.1 mV, QRS $\textgreater$ 100 ms \cite{rautaharju2009aha}):
\begin{equation}
P_{\text{MI}} = \min(0.9, 0.1 + \min(0.6, 3 \cdot |ST|) + 0.2 \cdot(\Delta_{\text{QRS}} > 100))
\label{eq:pmi}
\end{equation}
For QT abnormalities ($\textgreater$460 ms or $\textless$350 ms \cite{tisdale2020drug}):
\begin{equation}
P_{\text{QT}} = \min(0.9, 0.6 + \frac{\Delta_{\text{QT}} - 460}{100})
\label{eq:pqt}
\end{equation}
For hypertrophy (R $\textgreater$ 2.0 mV, QRS $\textgreater$ 110 ms \cite{cesini2018basic}):
\begin{equation}
\begin{aligned}
P_{\text{HE}} &= \min\Big(0.9,\, 0.1 + \min\big(0.5, \tfrac{A_R - 2.0}{2.0}\big) \\
&\quad + \min\big(0.3, \tfrac{\Delta_{\text{QRS}} - 110}{30}\big)\Big)
\end{aligned}
\label{eq:phyp}
\end{equation}
The disease with the highest probability is picked, with matches determined by priority. Myocardial Infarction $\textgreater$ Arrhythmias $\textgreater$ Conduction Disorders $\textgreater$ QT Abnormalities $\textgreater$ Hypertrophy, in keeping with clinical urgency \citep{yancy20182017}. Visualizations, such as annotated 2D waveforms and 3D probability bars enhance clarity.

\section{EXPERIMENTS}
We provide details the experimental setup used to evaluate the performance of FoundationalECGNet.

The experiments were conducted on a single NVIDIA RTX 4070 GPU to ensure efficient training, validation and testing. The model was trained using BCEWithLogitsLoss as the loss function. Training incorporated early stopping and a weight decay of 1e-4, employing the Adam Optimizer with a learning rate of 0.0005. During training, an early stopping mechanism was applied with a patience threshold of 10, monitoring validation accuracy in maximum mode. The overall process of training the proposed network and automatic ECG classification has been described in Algorithm \ref{algo}.
\begin{table*}[htbp]
\centering
\caption{STATISTICS OF EMPLOYED DATASETS}
\label{table:datasets}

\begin{tabular}{c c c c c}
\hline
\textbf{Reference} & \textbf{Dataset} & \textbf{Sample Size} & \textbf{Frequency (Hz)} & \textbf{Usage} \\
\hline
\cite{PTB-XL} & PTB-XL & 21,799 & 500 & Training \\
\cite{AF} & CinC 2017 & 8,528 & 300 & Training \\
\cite{MedalCare-XL} & MedalCare-XL & 16,900 & 500 & Training \\
\cite{PTB-Diagnos} & PTB & 549 & 1000 & Fine-tune and Testing \\
\hline
\end{tabular}

\end{table*}

Table \ref{table:datasets} presents all the datasets involved in this process. We apply our proposed model to four distinct and publicly available datasets to evaluate its performance. We resample all signals to 250 Hz. To ensure uniformity across the datasets, each signal is resampled to 1500 samples from each channel. To address class imbalance within the dataset, we employ the ADASYN algorithm, which generates synthetic samples for the underrepresented classes. The first three datasets are split into training and validation sets with an 80:20 ratio, and the final dataset is split into training and testing sets with an 80:20 ratio to evaluate the model's performance. During the training session, we trained one dataset at a time and transferred the updated weights to the following dataset, allowing it to learn from the different datasets gradually.

\section{RESULTS AND DISCUSSIONS}
\subsection{Evaluation Metrics}
The performance of our models are evaluated using complete range of evaluation metrics. They are accuracy (Acc), area under the curve (AUC), precision, recall, F1-score, and specificity [20]. Accuracy is the ratio of the correctly predicted instances to the total predicted instances, precision is the ratio of the correctly predicted positive observations to the total predicted positive observations, recall (also called sensitivity) quantifies the number of correct positive results by the number of all positive results, the F1-score is the harmonic mean of precision and recall, which gives a balanced performance, especially in cases of imbalance in the classes \citep{powers2020evaluation}. AUC reflects the ability of the model to discriminate between classes at all thresholds and a high value correlates with better discrimination \citep{huang2005using}. Specificity is the true negatives detected among all the actual negatives \citep{lalkhen2008clinical}. These metrics were calculated for binary classification (Normal vs. Abnormal ECG) and for multiclass classification (specific disease categories: AM, CD, MI, QT, and HE).

\subsection{Performance Evaluation}
We test two primary models: FoundationalECGNet (our proposed) and ExBlock+CBAM (\cite{exblock} adaptation from EEG to ECG data) on the PTB dataset, which consists of 15-channel ECG signals resampled to 250 Hz. The models are trained by three datasets and testing by one dataset. Results are presented for binary and multi-class tasks, highlighting differences between the models.

\begin{figure}[htbp]
\centering
\includegraphics[width=\linewidth]{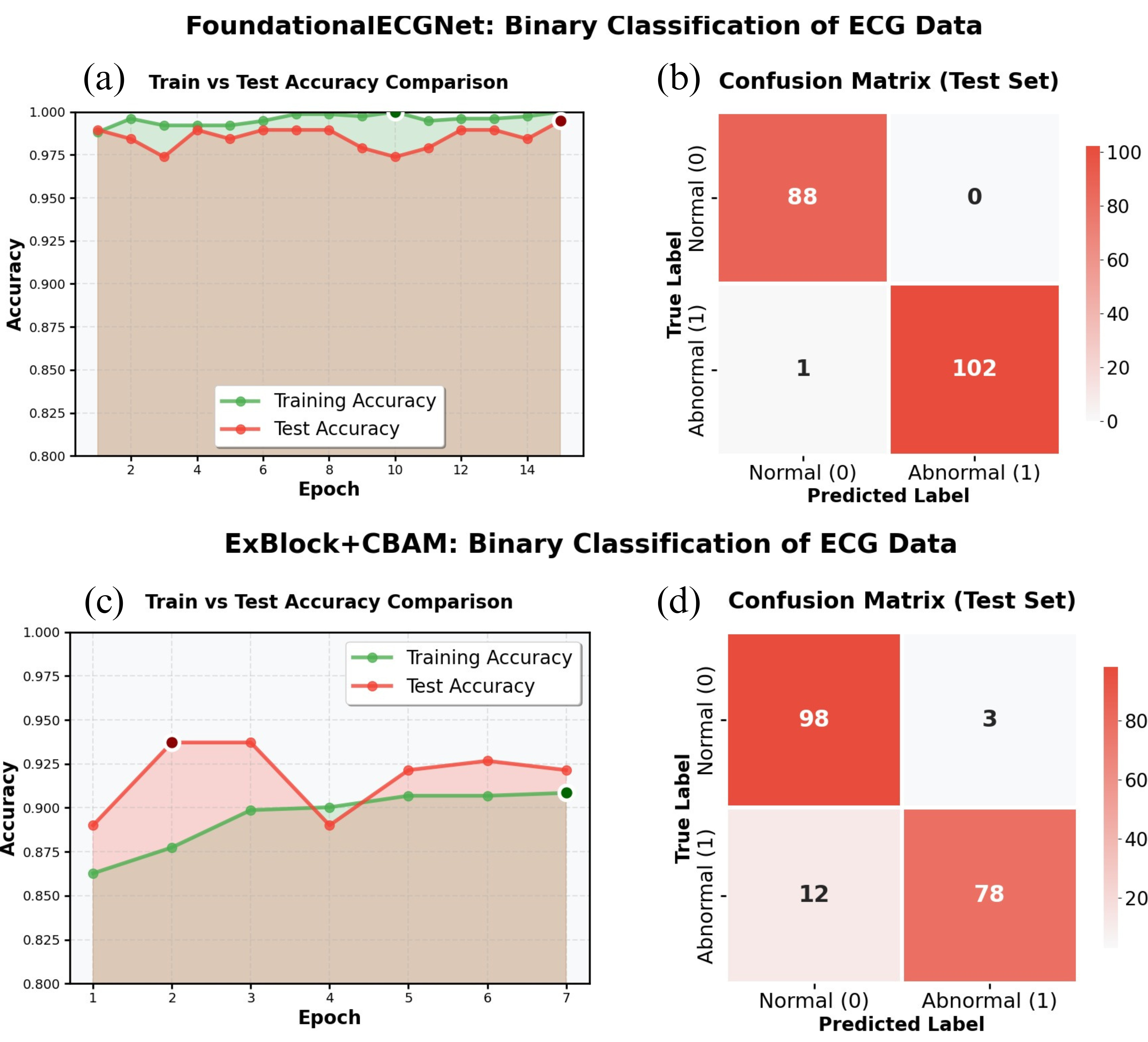}
\caption{Two models' performance for binary classification of ECG data. (a) and (c) show the comparison of training and test accuracy across epochs for FoundationalECGNet and ExBlock+CBAM, respectively. (b) and (d) represent the confusion matrices for the test set, corresponding to FoundationalECGNet and ExBlock+CBAM, respectively, with true labels categorized as Normal (0) and Abnormal (1) against predicted labels.}
\label{fig:ExB_curve}
\end{figure}

\begin{figure}[htbp]
\centering
\includegraphics[width=\linewidth]{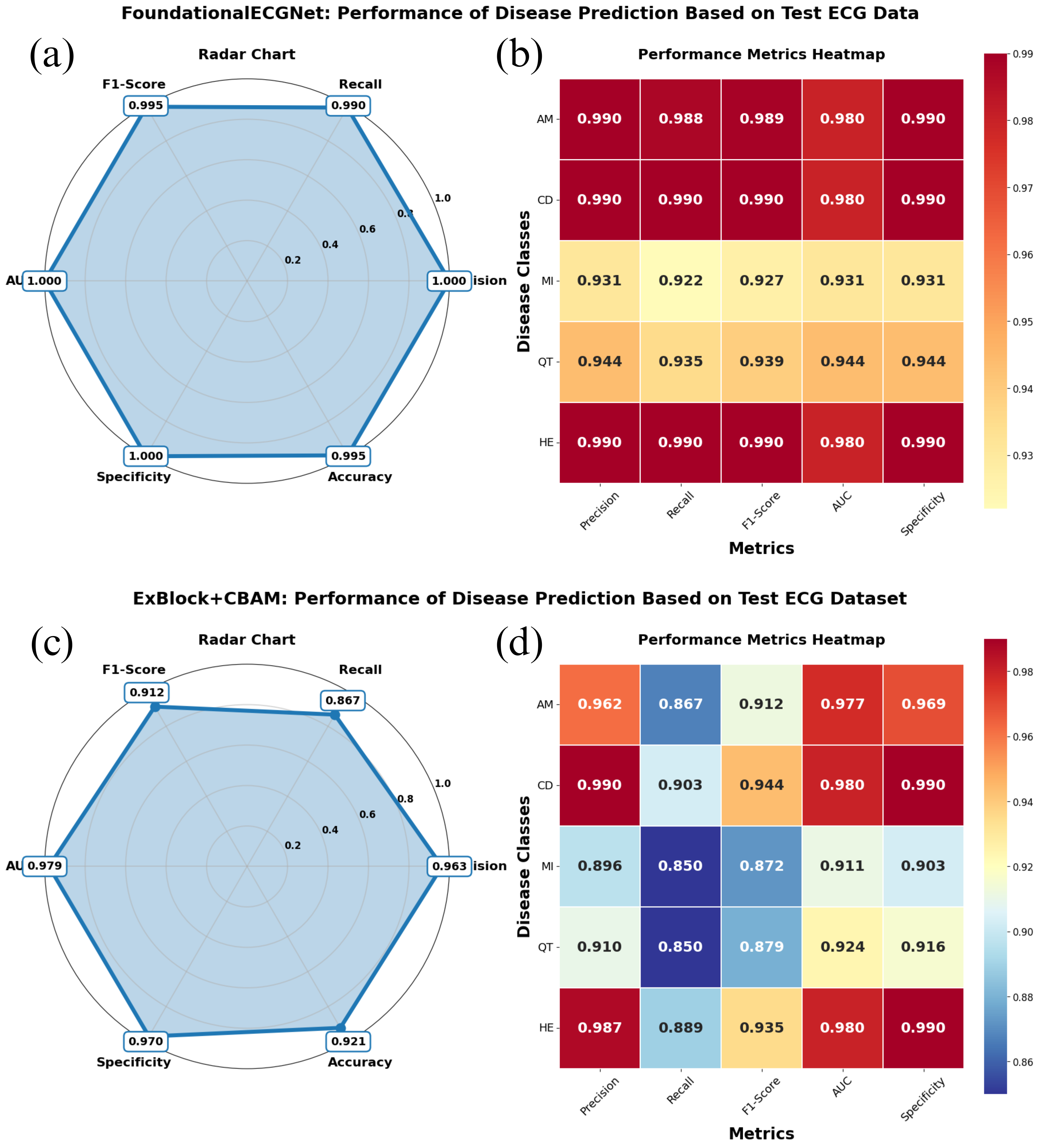}
\caption{Performance evaluation of two models for disease prediction based on test ECG data. (a) and (c) show the radar charts performance metrics including accuracy, specificity, F1-score, recall, and AUC for FoundationalECGNet and ExBlock+CBAM respectively. (b) and (d) present the performance metrics heatmaps, corresponding to FoundationalECGNet and ExBlock+CBAM respectively, with values for different disease classes (AM, CD, MI, QT, HE) across precision, recall, F1-score, AUC, and specificity.}
\label{fig:roc_curve}
\end{figure}

The performance difference between the baseline ExBlock+CBAM model and the suggested FoundationalECGNet is shown in Table \ref{tab:multi_class_metrics} and Fig. \ref{fig:roc_curve}. In the binary classification task, FoundationalECGNet achieved an accuracy of 99.48\%, above the baseline's 94.23\%, resulting in an absolute improvement of 5.25\%. This improvement is seen in the confusion matrices of Fig. 2, where the baseline model generates a greater number of false negatives, whereas our model detects abnormal ECGs with significantly better consistency. The training and test accuracy curves in Fig. \ref{fig:ExB_curve} show that FoundationalECGNet advances more rapidly and stabilizes at a superior performance level, showing an enhanced capacity for generalization beyond the training data. This is especially critical in clinical screening, because even a minor percentage of overlooked abnormal cases might result in serious issues. 

In the multiclass classification of abnormal ECG categories, FoundationalECGNet consistently exhibits its superiority. The macro-averaged F1-score increases from 0.913 for ExBlock+CBAM to 0.967 for FoundationalECGNet, exhibiting an overall improvement of 5.4\%. The improvements achieved across specific disease categories demonstrate the effectiveness of addressing the most complex patterns. The F1-score for myocardial infarction rises from 0.884 to 0.926, while for QT abnormalities it improves from 0.890 to 0.939. In arrhythmia classification, the improvement is notable, with the F1-score increasing from 0.919 to 0.988 and recall rising from 0.859 to 0.990. Our model achieves F1-scores of 0.990 for conduction abnormalities and hypertrophy, in contrast to the baseline scores of 0.940 and 0.932, respectively. The constant improvements indicate that FoundationalECGNet not only performs well on simpler scenarios but also identifies the complicated, irregular patterns often misclassified by less complex networks.

The baseline model's lower results, especially in myocardial infarction and QT anomalies, result from its limited capacity to integrate information over extended timeframes and across several leads. The dependence on convolutional and spatial attention layers limits its capability to identify localized features, performing adequately for conditions with clear, defined waveforms like hypertrophy or conduction disorders, but inadequately addressing the more complicated, temporally extended alterations associated with myocardial infarction and QT interval abnormalities. FoundationalECGNet addresses these drawbacks by incorporating dual-stage wavelet denoising alongside ExBlock, graph attention, and time-series transformer modules, thereby maintaining both noise-free shape and extensive temporal context. Consequently, it generates accurate and dependable predictions across all abnormal categories. 

In Figure \ref{fig:roc_curve}, the consistency of our model's performance. The radar chart for FoundationalECGNet provides comprehensive coverage across accuracy, recall, specificity, and AUC, while the baseline chart exhibits inconsistent coverage and notable shortcomings, especially in recall for myocardial infarction and QT abnormalities. The heatmaps indicate higher and more uniform metric values across all classes in our model, as opposed to the scattered and lower values of the baseline. Additionally, compared to the baseline, FoundationalECGNet reduces over half of the missed detections, as shown by the recall improvements in arrhythmia and conduction abnormalities. This indicates the combined effect of balanced training with ADASYN and the implementation of attention mechanisms, enabling the model to highlight rare yet clinically pertinent patterns during training.

\begin{figure*}[htbp]
\centering
\includegraphics[width=0.7\textwidth, height=10cm]{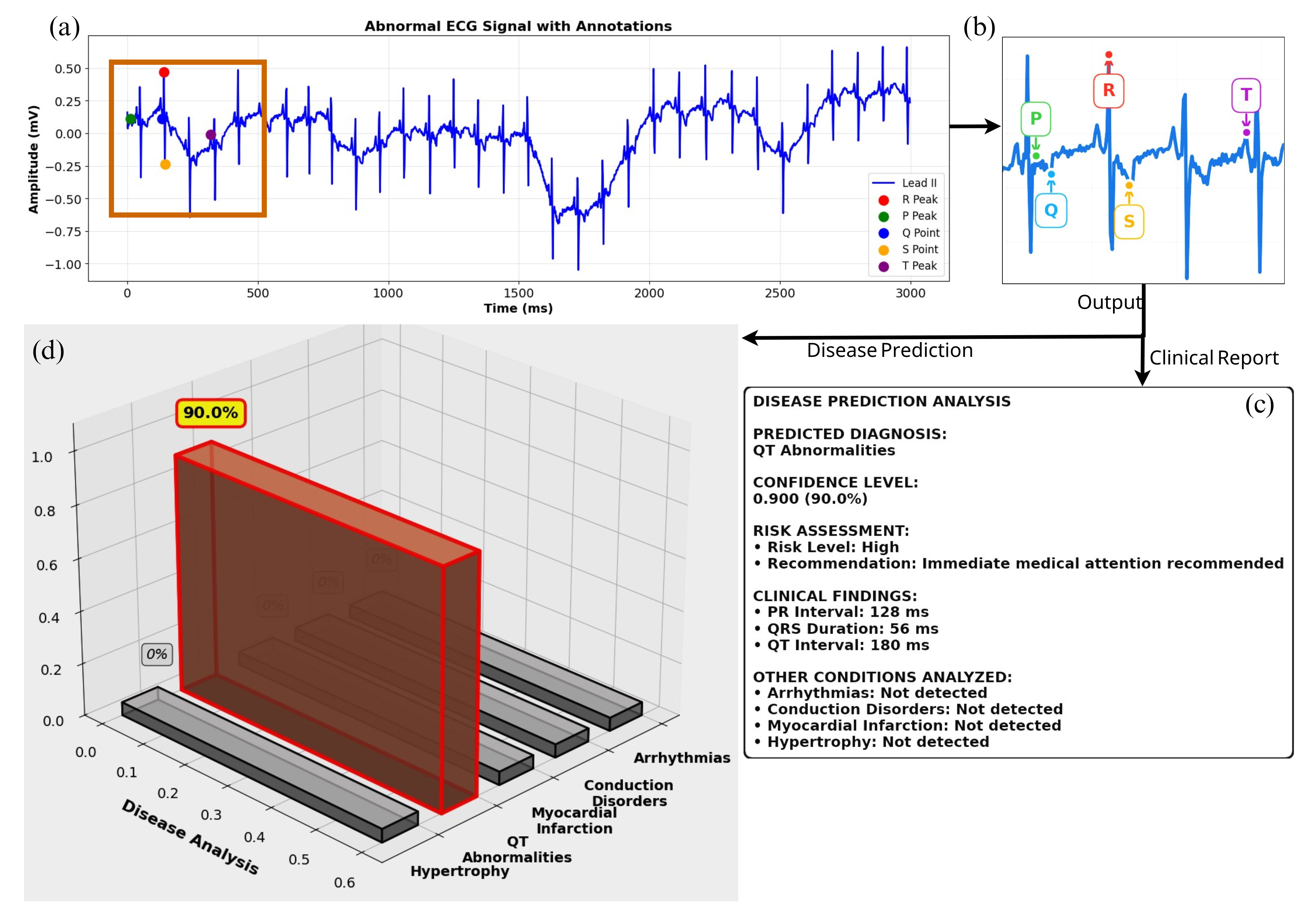}
\caption{Analysis of abnormal ECG signals and disease prediction outcomes. (a) Shows an abnormal ECG waveform with annotations. (b) Feature detection of P, Q, R, S, and T across a time series. (c) Presents a detailed analysis of the disease prediction, that contains the predicted diagnoses, confidence score, the risk factor, the clinical findings (PR interval, QRS duration) and other conditions considered. (d) Shows that the 3D confidence level can be used for disease analysis: the abnormality was detected with a confidence level.}
\label{fig:PS}
\end{figure*}

Our findings are further enhanced by Fig. \ref{fig:PS}, which depicts how the model's predictions can be seen and checked. Rather than only providing a disease label, the model additionally defines the significant features identified on the ECG, the metrics employed, and its confidence level regarding the diagnosis. This enables physicians to quickly examine the reasons behind the model's diagnosis and determine its reliability. The figure demonstrates that the model is not only precise but also user-friendly for medical professionals in practical use.

\begin{table*}[htbp]
\centering
\caption{MULTI-CLASS METRICS FOR ABNORMALITY SUBTYPES ON PTB ABNORMAL TEST SUBSET}
\label{tab:multi_class_metrics}
\begin{tabular}{@{}llccccc@{}}
\toprule
Class & Model & Precision & Recall & F1-Score & AUC & Specificity \\ \midrule
AM & ExBlock+CBAM & 0.962 & 0.867 & 0.912 & 0.977 & 0.969 \\
   & FoundationalECGNet & \textbf{0.990} & \textbf{0.988} & \textbf{0.989} & \textbf{0.980} & \textbf{0.990} \\
CD & ExBlock+CBAM & 0.990 & 0.903 & 0.944 & 0.980 & 0.990 \\
   & FoundationalECGNet & 0.990 & \textbf{0.990} & \textbf{0.990} & 0.980 & 0.990 \\
MI & ExBlock+CBAM & 0.896 & 0.850 & 0.872 & 0.911 & 0.903 \\
   & FoundationalECGNet & \textbf{0.931} & \textbf{0.922} & \textbf{0.927} & \textbf{0.931} & \textbf{0.931} \\
QT & ExBlock+CBAM & 0.910 & 0.850 & 0.879 & 0.924 & 0.916 \\
   & FoundationalECGNet & \textbf{0.944} & \textbf{0.935} & \textbf{0.939} & \textbf{0.944} & \textbf{0.944} \\
HE & ExBlock+CBAM & 0.987 & 0.889 & 0.935 & 0.980 & 0.990 \\
   & FoundationalECGNet & \textbf{0.990} & \textbf{0.990} & \textbf{0.990} & 0.980 & 0.990 \\
Macro-Avg & ExBlock+CBAM & 0.963 & 0.867 & 0.912 & 0.979 & 0.970 \\
          & FoundationalECGNet & \textbf{1.000} & \textbf{0.990} & \textbf{0.995} & \textbf{1.000} & \textbf{1.000} \\ \bottomrule
\end{tabular}
\end{table*}

\subsection{Ablation Studies}
The results of our ablation studies are presented in Table \ref{tab:ablation}. The model achieves the maximum performance, with a score of 0.9948 across all measures. The elimination of TST reduces the F1-score to 0.9686, showing its significance in detecting long-range temporal relationships essential for complex abnormality identification. Omitting GAT reduces the F1-score to 0.9790, highlighting its significance in simulating lead interactions. In the absence of CBAM, performance decreases to 0.9686, hence validating its improvement of spatial attention. Excluding both wavelet stages (Morlet and DB4) results in 0.9633, highlighting their combined effect in noise reduction while maintaining signal integrity. Replacing ExBlock decreases the F1-score to 0.9423, hence showing its critical role in efficient feature extraction. The statistics demonstrate that the integrated architecture simultaneously improves both robustness and accuracy in ECG analysis.

\begin{table*}[]
\centering
\caption{PERFORMANCE METRICS FOR THE ABLATION STUDIES COMPRISING DIFFERENT COMBINATIONS}
\label{tab:ablation}
\begin{tabular}{@{}cccccc ccccc@{}}
\toprule
\multicolumn{6}{c}{Modules} &  & \multicolumn{4}{c}{Performance} \\ 
\cmidrule(r){1-6} \cmidrule(l){8-11}
ExBlock & Morlet & DB4 & CBAM & GAT & TST &  & Accuracy & Precision & Recall & F1-Score \\ 
\midrule
$\checkmark$ & $\checkmark$ & $\checkmark$ & $\checkmark$ & $\times$ & $\times$ &  & 0.9372 & 0.9402 & 0.9372 & 0.9370 \\
$\checkmark$ & $\checkmark$ & $\checkmark$ & $\times$     & $\checkmark$ & $\times$ &  & 0.9686 & 0.9705 & 0.9686 & 0.9686 \\
$\checkmark$ & $\checkmark$ & $\checkmark$ & $\times$     & $\times$ & $\checkmark$ &  & 0.9791 & 0.9799 & 0.9791 & 0.9790 \\
$\checkmark$ & $\checkmark$ & $\checkmark$ & $\checkmark$ & $\times$ & $\times$ &  & 0.9476 & 0.9524 & 0.9476 & 0.9474 \\
$\checkmark$ & $\checkmark$ & $\checkmark$ & $\times$     & $\checkmark$ & $\checkmark$ &  & 0.9529 & 0.9568 & 0.9529 & 0.9527 \\
$\checkmark$ & $\times$     & $\times$     & $\checkmark$ & $\checkmark$ & $\checkmark$ &  & 0.9634 & 0.9658 & 0.9634 & 0.9633 \\
$\times$     & $\checkmark$ & $\checkmark$ & $\checkmark$ & $\checkmark$ & $\checkmark$ &  & 0.9424 & 0.9435 & 0.9424 & 0.9423 \\
$\checkmark$ & $\checkmark$ & $\checkmark$ & $\checkmark$ & $\checkmark$ & $\checkmark$ &  & \textbf{0.9948} & \textbf{0.9948} & \textbf{0.9948} & \textbf{0.9948} \\
\bottomrule
\end{tabular}
\end{table*}

\subsection{Computational Efficiency}
The computational complexity of all foundational models is presented in Table \ref{table:computational complexity}. We list the parameter count, FLOPs, and model size for each model. The most compact networks, including FoundationalECGNet (proposed) and CardX, comprise approximately 7.5 million and 15 million parameters, respectively, but the largest network, KED, contains 163.7 million parameters. FLOPs have a considerable range, spanning from 0.201 billion to 14 billion. Our model comprises 7.5 million parameters. The model size is quantified by its storage footprint in megabytes, with our model obtaining the minimum size of 28.62 MB. Our model is lightweight and can be trained on a single GPU. Consequently, from a complexity standpoint, our approach is potentially suitable for future implementation on personal electronic devices, allowing learning and optimizing based on individual ECGs.
\begin{table}[htbp]
\centering
\caption{COMPUTATIONAL COMPLEXITY COMPARED WITH STATE-OF-THE-ART FOUNDATIONAL MODELS}
\begin{tabular}{@{}lccc@{}}
\toprule
Model Name & Params & FLOPs & Model Size\\ \midrule
ECG-FM \citep{ecg-fm} & 90 million & 14 B & $\sim$ 360 MB \\
KED \citep{ked} & 163.7 million & -- & 1.3 GB \\
CardX \citep{cardX} & 15 million & \textbf{0.201 B} & $\sim$ 60 MB \\
FoundationalECGNet (proposed) & \textbf{7.5 million} & 0.469 B & \textbf{28.62 MB}  \\ \bottomrule
\end{tabular}
\label{table:computational complexity}
\end{table}

\section{CONCLUSION}
In this study, we propose FoundationalECGNet, a foundational model for automated ECG analysis that integrates dual-stage wavelet denoising with ExBlock, Convolutional Block Attention Module (CBAM), Graph Attention Networks (GAT), and Time Series Transformers (TST) to jointly capture spatial, inter-lead, and long-range temporal dependencies. The model achieves state-of-the-art performance in both binary (Normal vs. Abnormal) and multiclass disease classification tasks by utilizing multiple publicly available datasets. This shows that the model can reliably detect complicated diseases like hypertrophy, myocardial infarction, conduction disorders, and QT abnormalities. The main positive aspect of FoundationalECGNet is its interpretability and risk-level evaluation, offering clinicians clear diagnostic indicators instead of unclear predictions. These attributes make it an essential decision-support tool in practical contexts, particularly in rural and resource-limited situations where prompt diagnosis can be crucial for survival. The model's compact dimensions and computational efficiency reinforce its suitability for implementation across various clinical platforms. In the future, our efforts will concentrate on enhancing the architectural elements to better capture nuanced inter-lead dependencies and irregular temporal patterns, while also investigating self-supervised pre-training to diminish dependence on labeled data and broaden diagnostic coverage beyond the existing five major categories. We enhance the model's accessibility by using advanced visualization tools, including layer-wise attribution maps and counterfactual explanations, to promote more collaboration between AI systems and medical experts. These initiatives provide FoundationalECGNet as a scalable, transparent, and morally sound foundation for advanced AI-driven cardiology.


\printbibliography
\end{document}